# High-Throughput Approach to Modeling Healthcare Costs Using Electronic Healthcare Records

Alexander K Taylor
Department of Computer Science
University of Wisconsin - Madison
Madison, WI
ataylor24@wisc.edu

Ross Kleiman
Department of Computer Science
University of Wisconsin - Madison
Madison, WI
rkleiman@cs.wisc.edu

David Page
Department of Biostatistics and
Medical Informatics
University of Wisconsin - Madison
Madison, WI
page@biostat.wisc.edu

Peggy Peissig
Department of Biomedical Informatics
Marshfield Clinic Research Foundation
Marshfield, WI
peissig.peggy@marshfieldresearch.org

Scott Hebbring
Center for Precision Medicine Research
Marshfield Clinic Research Foundation
Marshfield, WI
hebbring.scott@marshfieldresearch.org

*Abstract*— Accurate estimation of healthcare costs is crucial for healthcare systems to plan and effectively negotiate with insurance companies regarding the coverage of patient-care costs. Greater accuracy in estimating healthcare costs would provide mutual benefit for both health systems and the insurers that support these systems by better aligning payment models with patient-care costs. This study presents the results of a generalizable machine learning approach to predicting medical events built from 40 years of data from >860,000 patients pertaining to >6,700 prescription medications, courtesy of the Marshfield Clinic in Wisconsin. It was found that models built using this approach performed well when compared to similar studies predicting physician prescriptions of individual medications. In addition to providing a comprehensive predictive model for all drugs in a large healthcare system, the approach taken in this research benefits from potential applicability to a wide variety of other medical events.

## I. Introduction

The accurate estimation of healthcare costs is important for a variety of stakeholders, including but not limited to the patient, the health insurer, and the healthcare system with which they interact. There are various models that define the agreement between healthcare systems and their insurance provider and dictate which organization will bear the burden of cost for certain patient healthcare costs, including outcome-based care and value-based care [1]. Each model requires healthcare systems and insurers to estimate the costs that will be incurred by the healthcare system over various time windows. These estimates are used to decide which costs will be covered by the healthcare system and which costs will be covered by the health insurers for a particular window of time. More accurate estimations of healthcare costs would aid both sides in reaching a mutually beneficial agreement.

One approach to modeling healthcare costs involves training machine learning models on electronic health records (EHRs) to predict all patient health events. However, attempting to integrate models from different studies predicting the broad range of possible health events would introduce a litany of issues dealing with standardizing the various models and dealing with the differences and assumptions inherent to each dataset. Therefore, this research focused on the construction of a widely applicable approach to building accurate predictions over a variety of medical conditions that is flexible enough to be applied to different datasets with minimal intervention.

Prior research [2] was able to implement such an approach through the construction of a large-scale data pipeline for the prediction of all diagnoses (ICD-9 codes) for seven different time windows (1 month, 6 months, 2 years, 5 years, 10 years, 15 years, and 20 years in advance). This work expands that pipeline to include the prediction of all prescription medications on a case-by-case basis. This extension enables predictions on an individual level for each medication, which in turn can be used to provide a prediction of how many patients will receive each medication. Given the market cost of each drug, these predictions can empower more accurate estimates of healthcare costs pertaining to medications. Furthermore, the general applicability and flexibility of the pipeline built for this research will allow for future research to make accurate estimates of costs for all medical events.

## II. Methods

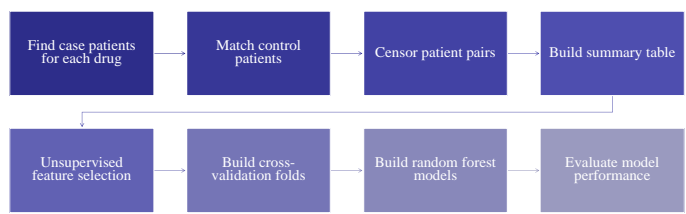

Fig. 1 Pipeline for the High-Throughput Prediction of Drug Prescriptions

### A. The pipeline

To build thousands of predictive models from raw EHRs, a pipeline was developed (Figure 1) that was general enough to guide the construction of accurate models for a wide variety of medical events. Using a pipeline format allowed for the creation of a structure that was both replicable in design and flexible in implementation, both of which are important features of an approach with a broad range of potential applications to EHRs.

One difficulty inherent to this task is providing a single method of formatting the EHRs used to train individual models to include the greatest amount of relevant data for thousands of unique medications, while preventing the prediction task from becoming artificially easy. Overcoming this obstacle requires proper handling and formatting of data to ensure that the models are not producing overly optimistic results. To address this challenge, two barriers were selected to creating an artificially easy task that would produce overly optimistic results: (1) control patient selection and (2) truncating patient pair data by windows of time.

This work utilized random forests [3] for model construction, as they are efficient to construct and perform well in high-dimensional settings such as those presented by EHR data. Unsupervised feature selection and cross-validation steps were added to the pipeline to evaluate the accuracy of the random forest models and the speed at which they were built. Importantly, the pipeline's modular approach allows for future implementations to substitute methods that would be better suited to their task with minimal friction.

### B. De-identification of patient data

The data used to conduct this research were obtained from the Marshfield Clinic, a healthcare system located in counties across northern, central, and western Wisconsin that currently provides care to more than 200,000 people.

Prior to sharing their EHRs, the Marshfield Clinic fully de-identified the protected health information of each patient and mapped each medical event to a unique, randomized numerical identifier. Continuous measurements, such as labs and vitals, were first mapped to a normalized reference range and were then de-identified. Mapping files were produced only for the names of the medical events (not protected information) and were used only to produce summary statistics and figures.

Prior to implementing the pipeline, a data cleaning process was performed on the raw EHR data. As this work aims to make predictions based on patient history, a per-patient data minimum was applied. Patients that did not meet a minimum of 4 total diagnoses and 2 unique visit dates were removed.

### C. Pre-implementation of pipeline

Medications can be hierarchically structured by their brand name, generic name, and drug class (e.g., Advil can be categorized as Advil, ibuprofen, and a non-steroidal anti-inflammatory drug). For this reason, predictions of drug prescriptions can be performed at differing levels of specificity, where brand name is the most specific and drug class is the least specific. The approach taken in this research focused on forming predictions at the generic level, which contains drugs that are unique based on their active ingredient but not necessarily by their manufacturer. This approach allows predictions for the widest possible range of drugs while maximizing the amount of data for each drug, thereby providing the most accurate possible estimate of the number of prescriptions over a given truncation window.

### D. Case-control identification and matching

The EHRs used in this research consisted of de-identified information for 862,947 patients that were collected over 40 years, including data pertaining to 6,712 unique medications. These data consisted of the de-identified date of the prescription, the Marshfield Clinic's unique patient identification code, and codes for the generic name, brand name, and drug class for each medication. After identifying all unique medications, lists were compiled for each medication that consisted of (1) all patients that were prescribed the medication at least once and (2) the date of each patient's first prescription. Restricting the prescription dates to each patient's first prescription prevented the prediction task from becoming artificially easy, since many medications treat chronic health conditions and the existence of prior prescriptions can itself indicate future prescriptions of the same medication.

To exclude datasets with insufficient training data, models were only constructed for those drugs that were prescribed to 500 or more patients; this restriction reduced the number of generic drugs modeled from 6,712 to 2,438.

For each medication, each case patient was matched to a control patient. The following criteria were used for control patients in order to create matching case-control pairs:
- No record of having been prescribed the medication being analyzed
- Same sex as the case patient
- Date of birth within 30 days of the case patient's
- Date of last known contact with the Marshfield Clinic was after the prescription was given to the case patient

These criteria were used to lessen the selective pressure of the age and sex variables, which may make the task artificially easy; the condition of 'last known contact' was added to verify that the control patient was still being seen by the Marshfield Clinic at the time the case patient received his or her prescription. Each case and control patient were used only one time in the selection process for each medication.

### E. Truncating the case-control pair data and building the summary table

To build models that can make accurate predictions of future prescriptions, it is necessary to truncate all data prior to the first prescription for the length of the time window being predicted. In other words, if the model is to predict a prescription 1 month in advance, all events recorded in the EHR for 30 days prior to the first prescription were removed from the training data. This step prevents the model's accuracy from becoming overly optimistic by removing medical events that are either a direct cause of the prescription or correlated with the direct cause, e.g., visiting the clinic and being diagnosed with a heart condition, which would result in the prediction task becoming artificially easy (Figure 2).

This data truncation was completed for all case-control patient data pertaining to each of the medications and was repeated for all time windows. The truncated case-control pair data were then assembled into a single relational summary table with one row per patient and each column representing a feature, creating the data format required for implementation of the random forest model.

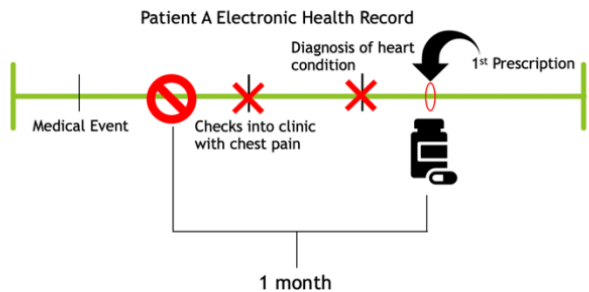

Fig. 2 Visual Representation of Data Truncation over a 1-Month Window

*F. Feature selection and building the random forest models*

Given the aim of performing high-throughput construction of many thousands of predictive models for a variety of generic drugs, random forest classifier models were chosen for their ability to retain accuracy while building predictions from high-dimensional data [3]–[4]. Indeed, EHRs contain large amounts of information for each patient, including diagnoses, prescriptions, labs, measurements of vitals, and other data, leading to a large volume of high-dimensional data. The random forest is an ensemble machine learning model that consists of an aggregation of smaller scale decision tree machine learning models. The implementation of the random forest used for this research was the scikit-learn 0.16.1 *RandomForestClassifier* model. The random forest also generalizes well to many different tasks while training quickly enough to accommodate the thousands of models built for this work [3].

The process of building the models began with setting hyperparameters that have performed well in previous studies in predicting medical events [2]. Namely, prior to building the random forest models, unsupervised feature selection was employed to retain features present in more than 1% of the population of both case and control patients. Each random forest model was then constructed from 500 trees that randomly selected 10% of the features identified in unsupervised feature selection as candidates for each split. Models were then evaluated using a 10-fold cross-validation scheme with the area under the receiver operating characteristic curve (AUC) used to measure performance.

*G. Condor High Throughput Computing*

This research was performed using the computer resources and assistance of the UW-Madison Center for High Throughput Computing (CHTC) in the Department of Computer Sciences. The CHTC is supported by UW-Madison, the Advanced Computing Initiative, the Wisconsin Alumni Research Foundation, the Wisconsin Institutes for Discovery, and the National Science Foundation, and is an active member of the Open Science Grid, which is supported by the National Science Foundation and the U.S. Department of Energy's Office of Science.

## III. RESULTS

The distribution of AUCs from each time window predicted was compared using kernel density estimation (KDE) across all time windows as a summary of model performance. Figure 3 shows that predictions over a longer time period had a lower AUC; the mean AUC ranged from $0.812 \pm 0.058$ at the 1-month truncation window to $0.666 \pm 0.050$ at the 5-year truncation window. It should be noted that fewer models were built at the 5-year truncation window (2,375) than at the 1-month truncation window (2,432) because some models did not meet the number of valid case-control pairs required for building a model. It was also observed that a group of high-performing models formed a separate distribution beginning at an AUC of 0.9167 for the 1-month models (Figure 3). Upon further investigation, it was found that this group of 172 models contained a variety of drugs that treat conditions that are easily predicted using recent bloodwork (e.g., diabetes).

## IV. DISCUSSION

The pipeline approach predicted many prescriptions with reasonable accuracy 1-month to 5 years in advance of initial prescription (Figure 3). These results were comparable to results from similar research predicting medical events at the 1-month time window [2]. For example, this previous study [10] predicted the prescription of trazodone over a 20-day truncation window with an AUC of 0.747; by comparison, the pipeline approach produced a model for the same medication with an AUC of 0.789 for a 1-month truncation window (30 days).

Unsurprisingly, analysis of the results revealed that accuracy decreases as the models predict a prescription further in advance. This fact may be due in part to certain medications being used to treat acute conditions; one such example is acute bronchitis, which relies heavily on medical events in the month leading up to the prescription of a cough suppressant.

Despite the decrease in AUC these models produce when predicting over a large truncation window, this research proposes that these models are still valuable in that they provide an alternative perspective to typical healthcare expenditure forecasts by utilizing EHRs to build predictions rather than extrapolating from past healthcare spending and economic trends [5].

The distribution found in the 1-month models above an AUC of 0.9167 contains medications whose models were performing artificially easy tasks that relied heavily on diagnoses near to the date of first prescription, such as a diagnosis of diabetes mellitus for insulin or chronic kidney disease for sevelamer. The gradual reduction in the prominence of this foothill in the distributions over the increasing truncation windows may be attributed to the censoring of these diagnoses as the censor window increased. The models for medications in this group would likely benefit from dynamic definition refinement (DDR) [2], which is a technique that further refines case-control patient selection criteria. DDR automatically determines a set of prerequisite diagnoses that both the case and control patient must have, which would reduce leakage and make the prediction task appropriately difficult. The addition of DDR to the generalized pipeline discussed in this research will be left for future research. To produce cost estimates of each drug, the only further step necessary is to compile a comprehensive list of all generic drug prices and use the predictive models to estimate the number of patients that will be prescribed each medication. Due to constraints on the accessibility of a list of all drug prices, this step will also be left for further research.

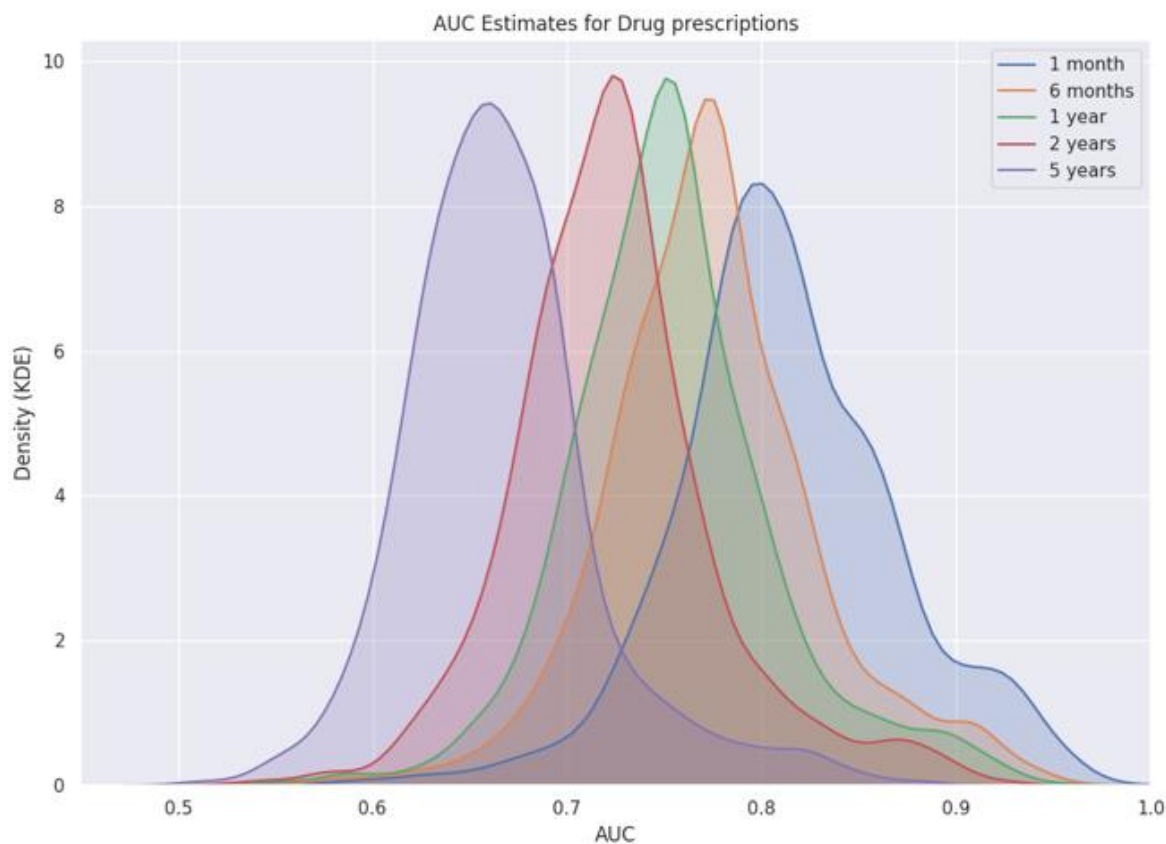

Fig. 3 Comparison of KDE of AUC curves across truncation windows

Notably, the only metric by which to judge the performance of our models is the performance of models from similar studies – no standard has been defined by which to measure the efficacy of such models. One such measure could be the accuracy of our models compared to a physician's ability to predict the prescribing of a given medication. Unfortunately, the lack of such a standard precludes us from being able to determine how effective the models built from the pipeline would be in an actual healthcare setting.

This study sought to establish the efficacy of building models for the high-throughput prediction of drug prescriptions. The results support the initial claim that a high-throughput approach to the prediction of prescriptions is not only viable but has applications in estimating healthcare costs with reasonable accuracy. In addition to supporting prior research, we believe our approach represents an important step in the incorporation of machine learning models into clinical decision support systems as well as a potential tool for large-scale investigation into determining previously unknown features that impact the prescriptions of all medications.